\documentclass[runningheads]{llncs}
\usepackage{times}
\usepackage{epsfig}
\usepackage{color}
\usepackage{xcolor}

\usepackage{algorithm}
\usepackage{algpseudocode}
\usepackage{threeparttable}
\usepackage{boldline}
\usepackage{epsfig} % for postscript graphics files
\usepackage{epstopdf} %converting to PDF\textbf{}
\usepackage{wrapfig}
\usepackage{kantlipsum}
\usepackage{graphicx}
\usepackage{amsmath,amssymb} % define this before the line numbering.

\usepackage{subcaption}
\usepackage{array}
\newcolumntype{P}[1]{>{\centering\arraybackslash}p{#1}}
\usepackage{xspace}
\definecolor{jscolornotes}{rgb}{0.2,0.6,0.2}
\definecolor{sicolor}{rgb}{0.1,0.1,0.6}

\newcommand{\jsremove}[1]{} % to remove JSnotes
\definecolor{xxawfcolor}{rgb}{0.6,0.2,0.2}

\newcommand{\IGNORE}[1]{{}}

\newcommand{\seclabel}[1]{\label{sec:\reflabel-#1}}
\newcommand{\secref}[2][\reflabel]{Section~\ref{sec:#1-#2}}

\newcommand{\eqlabel}[1]{\label{eq:\reflabel-#1}}
\renewcommand{\eqref}[2][\reflabel]{(\ref{eq:#1-#2})}

\newcommand{\figlabel}[2][\reflabel]{\label{fig:#1-#2}}
\newcommand{\figref}[2][\reflabel]{Fig.~\ref{fig:#1-#2}}

\newcommand{\tablelabel}[2][\reflabel]{\label{table:#1-#2}}
\newcommand{\tableref}[2][\reflabel]{Table~\ref{table:#1-#2}}

\newcommand{\reflabel}{dummy} % Dummy initial reflabel - use renewcommand...

\newcommand{\be}{\begin{equation}}
\newcommand{\ee}{\end{equation}}
\newcommand{\bea}{\begin{eqnarray}}
\newcommand{\eea}{\end{eqnarray}}
\newcommand{\beas}{\begin{eqnarray*}}
\newcommand{\eeas}{\end{eqnarray*}}

\newcommand{\ie}{i.e.~}

\newcommand{\etal}{et al.\xspace}

\newcommand{\cut}[1]{}

\newcommand{\gausscomp}{j} % trace
\newcommand{\gausscompNum}{J} % trace
\newcommand{\gauss}{\mathcal{N}} 

\usepackage[width=122mm,left=12mm,paperwidth=146mm,height=193mm,top=12mm,paperheight=217mm]{geometry}

\begin{document}
% \linenumbers
% \pagestyle{headings}
% \mainmatter

% \title{Occlusion-aware Hand Pose Estimation Using Hierarchical Mixture Density Network} 
% \titlerunning{Occlusion-aware Hand Pose Estimation Using Hierarchical Mixture Density Network}
% \authorrunning{Qi Ye, Tae-Kyun Kim}
% \author{Qi Ye, Tae-Kyun Kim}

% \institute{ Department of Electrical and Electronic Engineering,Imperial College London
% }
% \maketitle

\def\ECCV18SubNumber{***}  % Insert your submission number here
\title{Occlusion-aware Hand Pose Estimation Using Hierarchical Mixture Density Network}
\titlerunning{}
\author{Qi Ye, Tae-Kyun Kim}
\institute{Imperial College London, London, UK}
 \maketitle

% \begin{document}
% \newcommand*\samethanks[1][\value{footnote}]{\footnotemark[#1]}
% % \renewcommand\thelinenumber{\color[rgb]{0.2,0.5,0.8}\normalfont\sffamily\scriptsize\arabic{linenumber}\color[rgb]{0,0,0}}
% % \renewcommand\makeLineNumber {\hss\thelinenumber\ \hspace{6mm} \rlap{\hskip\textwidth\ \hspace{6.5mm}\thelinenumber}}
% % \linenumbers

\begin{abstract}
Learning and predicting the pose parameters of a 3D hand model given an image, such as locations of hand joints, is challenging due to large viewpoint changes and articulations, and severe self-occlusions exhibited particularly in egocentric views. Both feature learning and prediction modeling have been investigated to tackle the problem. Though effective, most existing discriminative methods yield a single deterministic estimation of target poses. Due to their single-value mapping intrinsic, they fail to adequately handle self-occlusion problems, where occluded joints present multiple modes. In this paper, we tackle the self-occlusion issue and provide a complete description of observed poses given an input depth image by a novel method called hierarchical mixture density networks (HMDN). The proposed method leverages the state-of-the-art hand pose estimators based on Convolutional Neural Networks to facilitate feature learning, while it models the multiple modes in a two-level hierarchy to reconcile single-valued and multi-valued mapping in its output. The whole framework with a mixture of two differentiable density functions is naturally end-to-end trainable. In the experiments, HMDN produces interpretable and diverse candidate samples, and significantly outperforms the state-of-the-art methods on two benchmarks with occlusions, and performs comparably on another benchmark free of occlusions. 
\end{abstract}

\section{Introduction}
\seclabel{sec:introduction}

3D hand pose estimation has shown an increasing interest with commercial miniaturized RGBD cameras and its ubiquitous applications in virtual/augmented reality (VR/AR) \cite{Jang_TVCG_2015}, sign language recognition \cite{Chang_CVIU_2016,Yin_ECCV_2016}, activity recognition \cite{Rohrbach_CVPR_2012}, and  man-machine interfaces for robots and autonomous vehicles. There are generally two typical camera settings: a third-person viewpoint, where the camera is set in front of the user, and an egocentric (or first-person) viewpoint, where the camera is mounted on the user's head (in VR glasses, for example), or shoulder. While both settings share challenges like the full range of 3D global rotations, complex articulations, self-similar parts of hands, self-occlusions are more dominant in the egocentric viewpoints. Most existing hand benchmarks are collected in the third-person viewpoints, e.g. the two widely used ICVL \cite{Tang_ICCV_2013} and NYU \cite{Tompson_TOG_2014} have less than 9\% occluded finger joints.

Discriminative methods (cf. generative model fitting) in hand pose estimation learn a mapping from an input image to pose parameters from a large training dataset, and have been very successful in the settings of third-person viewpoints. However, they fail to handle occlusions frequently encountered in egocentric viewpoints. They treat the mapping to be single-valued, not being aware of that an input image may have multiple pose hypotheses when occlusions occur. See \figref{mainfigure1} where an example image and its multiple pose labels from the BigHand dataset \cite{Yuan_CVPR_2017} are shown. 

\begin{figure}[t]
	\centering
	\includegraphics[trim=0mm 0mm 120mm 100mm, clip, width=\linewidth]{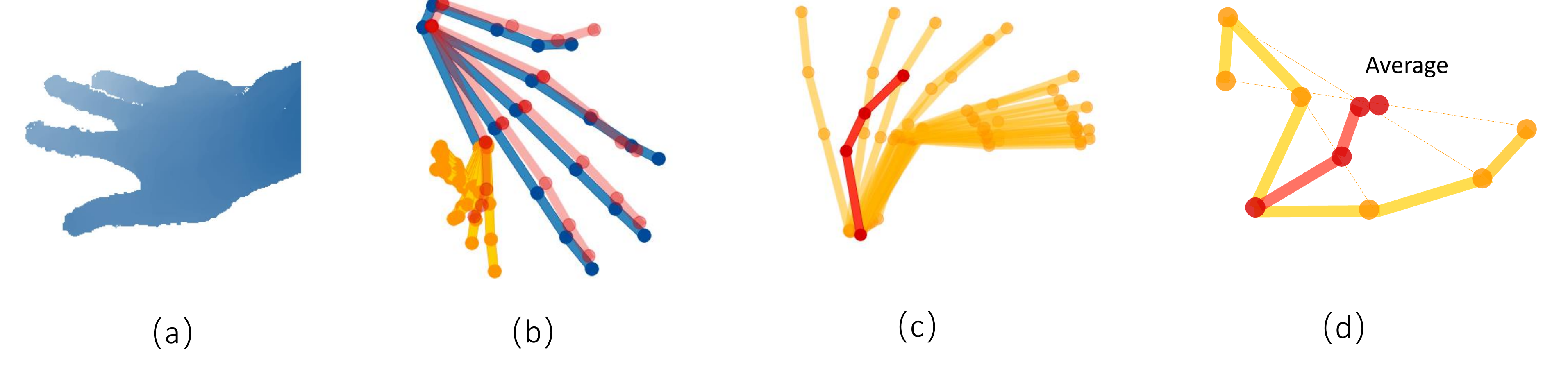}
    \caption{(a) A hand depth image with the pinky finger occluded. (b) Multiple pose labels (visible joints are in blue and occluded joints in yellow) and the predicted pose by CNN trained using a mean squared error (in red), in a 3D rotated view to better illustrate the problem (same for the following skeletons shown). (c) A closer look of the multiple labels and the CNN prediction on the occluded joints. (d) The average of two labels yields a physically implausible pose. }
	\figlabel{mainfigure1}
\end{figure}

Given a set of hand images and their pose labels i.e. 3D joint locations, discriminative methods such as Convolutional Neural Networks (CNN) minimize a mean squared error function, and the minimization of such error functions typically yields the averages of joint locations conditioned on input images. When all finger joints in the images are visible, the mapping is single-valued and the conditional average is correct, though the average only provides a limited description of the joint locations. However, for the occlusion cases, which happen frequently in the egocentric and hand-object interaction scenarios \cite{Oikonomidis_CVPR_12,Oikonomidis_ICCV_11,Poier_BMCV_2015,Garcia-Hernando_CVPR_2018}, the mapping is multi-valued due to occluded joints that exhibit multiple locations given the same images. The conditional average of the joint locations is not necessarily a correct pose, as shown in \figref{mainfigure1}b and \figref{mainfigure1}c. The prediction of a CNN trained by the mean squared error function is shown in red. It is interpretable and close to the ground truth for the visible joints, whereas it is physically implausible and not close to any of the given poses for the occluded joints. The example is clearer in \figref{mainfigure1}d, where we are given two available poses for the same image and CNN trained with the mean squared error function produce the pose estimation in red.

Existing discriminative methods, including the above CNN, are mostly deterministic, i.e. their outputs are single poses, thus lacking the description of all available joint locations. A discriminative method often serves as the initialization of a generative model fitting in the hybrid pose estimation approaches \cite{Tang_ICCV_2015,Sharp_CHI_2015}. If the discriminative method yields a probability distribution that well fits the data, than a single deterministic output, it would allow sampling pose hypotheses from its distribution.  This, in turn, reduces the search space helping a faster convergence and avoids local minima from diverse candidates in the model fitting. Such sampling is crucial also for multi-stage pose estimation \cite{Tang_ICCV_2015} and hand tracking \cite{Oikonomidis_BMVC_11}. Previous methods ignore the pose space to be explored ahead and their optimization frameworks are not aware of occlusions.

%can not treat the optimization of visible and occluded joints simultaneoulsy. 

% not only outputs an average, which is often physically implausible, but 
 
%For generative methods which search the entire pose space to find a best hypothesis that minimizes the discrepancy between the rendered image and the input image, 

\begin{figure*}[t]
\includegraphics[trim=10mm 50mm 10mm 50mm, clip, width=\linewidth]{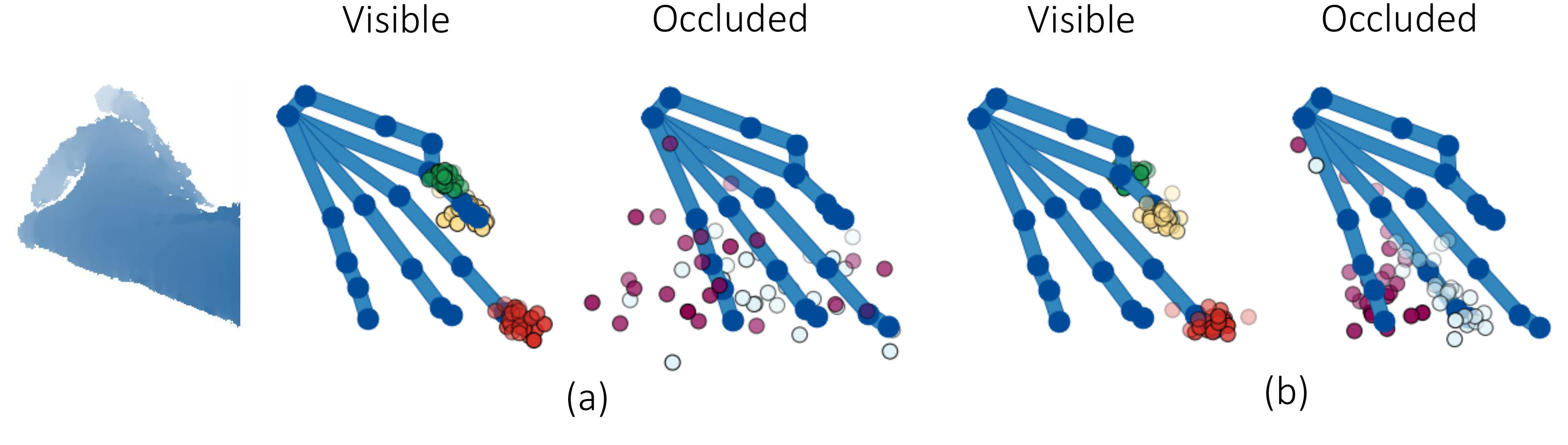}
\caption{Samples drawn from the distributions of (a) SGN and (b) HMDN for finger tips.}
\figlabel{qualiModels_brief}
\end{figure*}

%It models conditional probability distributions of the joint locations. 

% To leverage CNN for effective feature learning, 

In this paper, hierarchical mixture density networks (HMDN) are proposed to give a complete description of hand poses given images under occlusions. The probability distribution of joint locations is modeled in a two-level hierarchy to consider both single- and multi-valued mapping conditioned on the joint visibility. The first level represents the distribution of a latent variable for the joint visibility, while the second level the distribution of joint locations by a single Gaussian model for visible joints or a Gaussian mixture model for occluded joints. The hierarchical mixture density is topped upon the CNN output layer, and the whole network is trained end-to-end with the differentiable density functions. See \figref{qualiModels_brief}. The distribution of the proposed method HDMN captures diverse joint locations in a compact manner, compared to the network that learns a single Gaussian distribution (SGN). To the best of our knowledge, HMDN is the first solution that has its estimation in the form of a conditional probability distribution with the awareness of occlusions in 3D hand pose estimation. The experiments show that the proposed method significantly improves several baselines and state-of-the-art methods under occlusions given the same number of pose hypotheses.

%or samples.

%(incluing one, so to compare existing deterministic methods in a fair manner). 

% is a distribution for the joint location dependent on the visibility: 

% the visible joint represented 

% by a single Gaussian kernel corresponding its single-valued mapping trait 

% and the occluded joint represented by a mixture of multiple Gaussian kernels to fit the multi-valued mapping.

\section{Related Work}

\subsection{Pose estimation under occlusion}

For free hand motions, methods explicitly tackling self-occlusions are rare as most existing datasets are collected in third-person viewpoints and the proportion of occluded joints is small. Franziska \etal \cite{Franziska_ICCV2017} observed that many existing methods fail to work under occlusions and even some commercial systems claiming for egocentric viewpoints often fail under severe occlusions. Methods developed for hand-object interactions  \cite{Oikonomidis_ICCV_11,Dimitrios_IJCV_2016,Sridhar_ECCV_2016}, where occlusions happen frequently, model hands and objects together to resolve the occlusion issues. Jang \etal \cite{Jang_TVCG_2015} and Rogez \etal \cite{Rogez_ECCVW_2014} exploit pose piors to refine the estimations. Franziska \etal \cite{Franziska_ICCV2017} and  Rogez \etal \cite{Rogez_CVPR_2015} generate synthetic images to train discriminative methods for difficult egocentric views.

In human body pose estimation and object keypoint detection, occlusions are tackled more explicitly \cite{Haque_ECCV_2016,Charles_CVPR_2016,Rafi_ICCVW_2015,Ghiasi_CVPR_2014,Sigal_CVPR_2006,Chen_CVPR_2015,Hsiao_CVPR_2012,Navaratnam_ICCV_2007}. 
Chen \etal \cite{Chen_CVPR_2015} and Ghiasi \etal \cite{Ghiasi_CVPR_2014} learn templates for occluded parts. Hsiao \etal \cite{Hsiao_CVPR_2012} construct a occlusion model to score the plausibility of occluded regions. Rafi \etal \cite{Rafi_ICCVW_2015} and Wang \etal \cite{WangHough_CVPR_2013} utilize the information in backgrounds to help localize occluded keypoints. 
Charles \etal \cite{Charles_CVPR_2016} evaluate automatic labeling according to occlusion reasoning. Haque \etal \cite{Haque_ECCV_2016} jointly refine the prediction for visible parts and visibility mask in stages. Navaratnam \etal \cite{Navaratnam_ICCV_2007} tackle the multi-valued mapping for 3d human body pose via marginal distributions which help estimate the joint density.

The existing methods do not address multi-modalities nor do not model the difference in distributions of visible and occluded joints. For CNN-based hand pose regression \cite{Oberweger_ICCV_2015,Oberweger_ICCVW_2017,Tompson_TOG_2014,Ye_ECCV_2016}, the loss function used is the mean squared error, bringing in the aforementioned issues under occlusions. For random forest-based pose regression \cite{Tang_ICCV_2013,Sun_CVPR_2015,Sharp_CHI_2015}, the estimation is made from the data in leaf nodes and it is convenient to fit a multi-modal model to the data. However, with no information of which joints are visible or occluded, the data in all leaf nodes is captured either by the mean-shift (a uni-modal distribution) or a Gaussian Mixture Model (GMM)  \cite{Tang_ICCV_2015}. 

\subsection{Mixture Models}
Mixture density networks (MDN) were first proposed in \cite{bishop1994mixture} to enable neural networks to overcome the limitation of the mean squared error function by producing a probability distribution. Zen \etal~\cite{Zen_ICASSP_2014} use MDN for acoustic modeling and Kinoshita \etal~\cite{Kinoshita_ICASSP_2017} for speech feature enhancement. Variani \cite{Variani_ICASSP_2015} proposes to learn the features and the GMM model jointly. All these work apply MDN to model acoustic signals without an adaptation of the mixture density model. Our paper extends MDN by a two-level hierarchy to fit the specific mixture of single-valued and multi-valued problems, for the application of hand pose estimation under occlusions. To model data under noise, a similar hierarchical mixture model is proposed in \cite{Constantinopoulos_PAMI_2006} to represent ``useful" data and ``noise" by different sub-components, and a Bayesian approach is used to learn the parameters of the mixture model. Different from the work, we model a conditional distribution and use CNN to discriminatively learn the model parameters.

\section{Hierarchical Mixture Density Network}

\subsection{Model Representation}
\seclabel{sec:hmdn}
%\vspace{-10mm}
\begin{figure}[t]
	\centering
	\includegraphics[trim=0mm 0mm 75mm 120mm, clip, width=\linewidth]{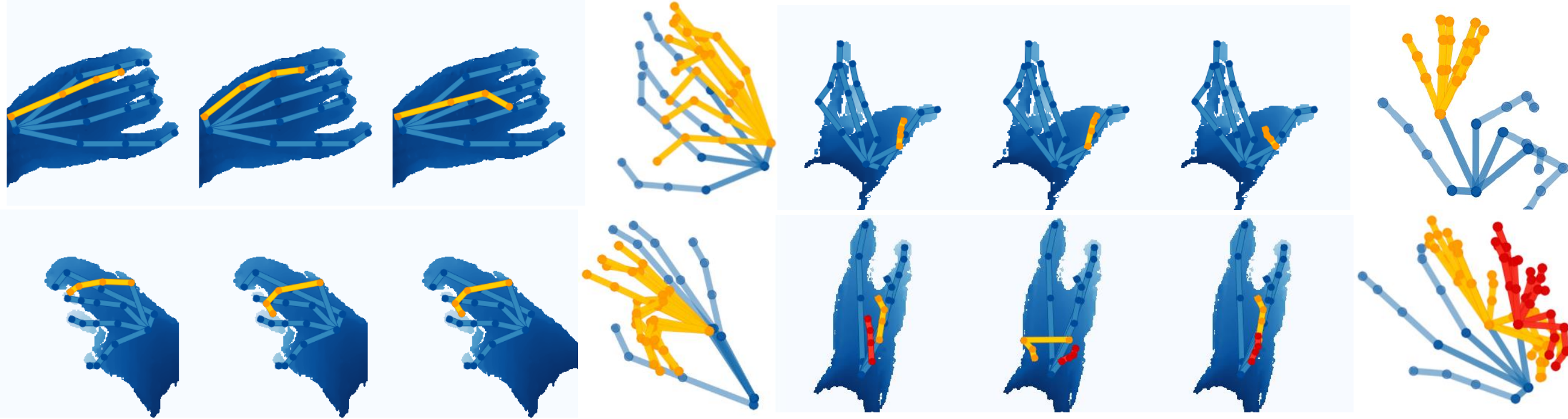}	
	\caption{Hand images under self-occlusions exhibiting multiple pose labels. Each shows different example labels overlaid on the same depth image (in the first three columns), and all available labels in a 3D rotated view (in the last column).Visible joints are in blue and occluded joints in other colors. 
    }
	\figlabel{ImgMultGT}
\end{figure}

The dataset to learn the model consists of $\{x_n, Y_n^d, v_n^d | n=1,...,N, d=1,...,D \}$, where $x_n$, $Y_n^d$, and $v_n^d$ denote the $n$-th hand depth image, the multiple pose labels i.e. 3D locations of the $d$-th joint of the $n$-th image, and the visibility label of the $d$-th joint of the $n$-th image, respectively. The $d$-th joint, when it is occluded, is associated with multiple labels $Y_n^d=\{y_{nm}^{d}\}$, where $y_{nm}^{d}\in R^3$ is the $m$-th label i.e. 3D location. The number of labels varies and is one when the joint is visible. See \figref{ImgMultGT} for examples. The visibility label is binary, indicating whether the $d$-th joint of the $n$-th image is visible or not. We treat $D$ joints independently.

% As $d$ hand joints are independent, to facilitate the explanation, for all the variables below we leave the subscript $\handpart$. The distribution of $\target$ is conditioned on $\inputimg$, but for simplicity, we omit the condition $\inputimg$ in the following.

To model hand poses under occlusions, a two-level hierarchy is considered. The top-level takes the visibility label, and the bottom-level switches between a uni-modal distribution and a multi-modal distribution, depending on the joint visibility. 

The binary label or variable $v_n^d$ follows the Bernoulli distribution, 
\be
p(v_n^d|w_n^d) = (w_n^d)^{v_n^d}(1-w_n^d)^{(1-v_n^d)},
\eqlabel{visBernoulli}
\ee
where $w_n^d$ is the probability that the joint is visible. As existing hand benchmarks do not provide the joint visibility labels, we use a sphere model similar to \cite{Chen_CVPR_2014} to generate the visibility labels from the available pose labels. The sphere centers are obtained from the joint locations and depth image pixels are assigned to the nearest spheres. Hand joints whose spheres have the number of pixels below a threshold are labeled as occluded. See \figref{vislabel}. The visibility labels $v_n^d$ are used for training, and they are inferred at testing.

% Since existing hand benchmarks do not provide the joint visibility label, we use a sphere model similar to \cite{Chen_CVPR_2014}. The sphere centres are obtained from the pose lables, i.e. joint locations, and depth image pixels are assigned to the nearest spheres. Hand joints whose spheres have the number of pixels below a threshold are labelled as occluded. See \figref{vislabel}. The visibility labels $v_n^d$ are given for training, and they are inferred at testing.

When $v_n^d=1$, the joint is visible in the image and the location is deterministic. Considering the label noise, $y_{nm}^{d}$ is generated from a single Gaussian distribution, 
\be
p(y_{nm}^{d}|v_n^d=1) = \gauss(y_{nm}^{d};\mu_n^d,\sigma_n^d).
\eqlabel{uni}
\ee
When the joint is occluded i.e. $v_n^d=0$, it has multiple labels and they are drawn from a Gaussian Mixture Model (GMM) with $J$ components, 
\be
p(y_{nm}^{d}| v_n^d=0) = \prod\limits_{\gausscomp=1}^{\gausscompNum}\gauss(y_{nm}^{d};\epsilon_{nj}^d,s_{nj}^d)^{z_{nj}^d},
\eqlabel{gmm}
\ee
where $\epsilon_{nj}^d$ and $s_{nj}^d$ represent the center and standard deviation of the $j$-th component. A hidden variable $z_{nj}^d$ is in 1-of-$\gausscompNum$ representation. If $z_{nj}^d=1$, the joint location is drawn from the $\gausscomp$-th component. 
% $\latent_\gausscomp$ is one element of the $\gausscompNum$-dimensional latent variable $\latent$ and the variable is a 1-of-$\gausscompNum$ representation.
Assume the hidden variable is under the distribution 
$p(z_{nj}^d)=\prod\limits_{\gausscomp=1}^{\gausscompNum} 
(\pi_{nj}^d)^{(z_{nj}^d)}$, 
where $0 \leq \pi_{nj}^d \leq 1$, $\sum\limits_{\gausscomp=1}^{\gausscompNum}
\pi_{nj}^d = 1$.
%and $\cmixcoeff$ is short for $\{\mixcoeff_\gausscomp|\gausscomp=1,...,\gausscompNum\}$.
Eqn. \eqref{gmm} can be re-written as 
$
p(y_{nm}^{d}| v_n^d=0) = \sum\limits_{\gausscomp=1}^{\gausscompNum} \pi_{nj}^d \gauss(y_{nm}^{d};\epsilon_{nj}^d,s_{nj}^d)
$.
% $p(\target|\cmixcoeff, \cmeano,\cvaro,\vistarget=0) = \sum\limits_{\gausscomp=1}^{\gausscompNum} \mixcoeff_\gausscomp \gauss(\target;\meano_\gausscomp,\varo_\gausscomp)$.

%\vspace{-5mm}

\begin{figure}[t]
\centering
	\includegraphics[trim=40mm 65mm 60mm 60mm, clip, width=\linewidth]{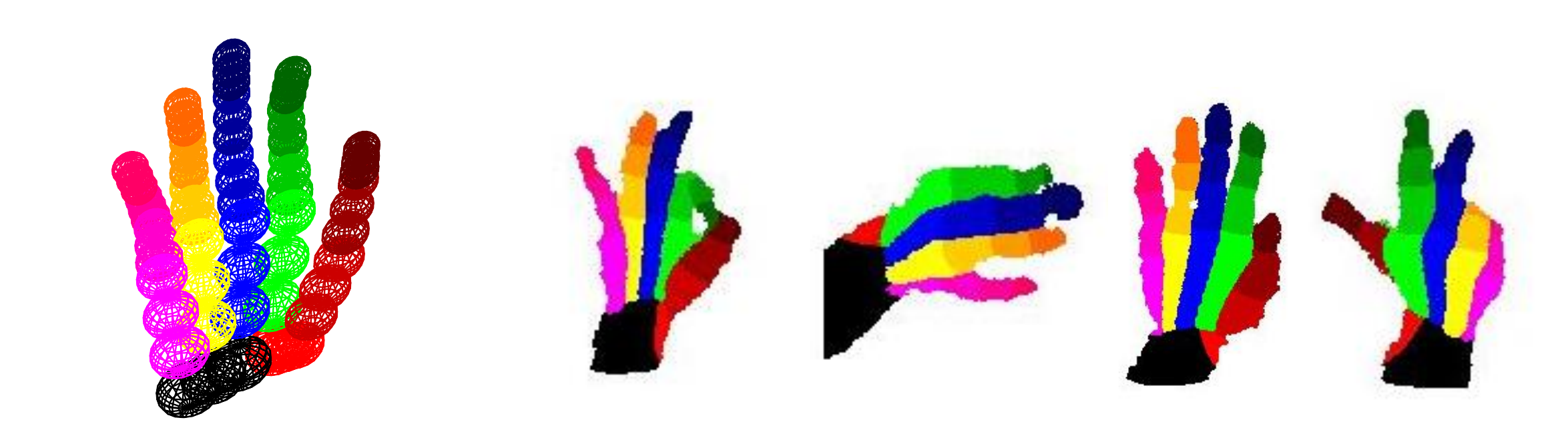}
\caption{Leftmost: the hand sphere model; Right: examples with pixels assigned to different parts}
\figlabel{vislabel}
\end{figure}

With all components defined, the distribution of the joint location conditioned on the visibility is 
\begin{equation}
p(y_{nm}^{d}|v_n^d) =\left[\gauss(y_{nm}^{d};\mu_n^d,\sigma_n^d)\right]^{v_n^d}\left[ \sum\limits_{\gausscomp=1}^{\gausscompNum} \pi_{nj}^d\gauss(y_{nm}^{d};\epsilon_{nj}^d,s_{nj}^d)\right] ^{(1-v_n^d)}
\eqlabel{yDistribution}
\end{equation}
and the joint distribution of $y_{nm}^{d}$ and  $v_n^d$ is
\begin{equation}
p(y_{nm}^{d},v_n^d) = \left[w_n^d \gauss(y_{nm}^{d};\mu_n^d,\sigma_n^d)\right]^{v_n^d}\left[(1-w_n^d) \sum\limits_{\gausscomp=1}^{\gausscompNum} \pi_{nj}^d\gauss(y_{nm}^{d};\epsilon_{nj}^d,s_{nj}^d)\right] ^{(1-v_n^d)}
\eqlabel{jointDistribution}
\end{equation}

Eqn. \eqref{yDistribution} shows that the generation of joint locations $y_{nm}^{d}$ given the input image $x_n$ is in a two-level hierarchy: first, a sample $v_n^d$ is drawn from Eqn. \eqref{visBernoulli} and then, depending on $v_n^d$, a joint location is drawn either from a uni-modal Gaussian distribution or GMM. Thus, the proposed model switches between the two cases and provides a full description of hand poses under occlusions. The joint distribution in Eqn. \eqref{jointDistribution} is used to define the loss function in \secref{sec:traintest}.

\subsection{Architecture}
\seclabel{sec:archi}

% \target_\handpart^\gtIdx,\vistarget_\handpart|\visbleProb_\handpart,\meanv_\handpart,\varv_\handpart,\{\meano_{\handpart\gausscomp}\},\{\varo_{\handpart\gausscomp}\},\{\mixcoeff_{\handpart\gausscomp}\}) $.

The formulations in the previous section are presented for the $d$-th joint $y_{nm}^{d}$. For all $D$ joints of hands, the distribution is obtained by multiplying the distributions of independent joints. The observed hand poses and the joint visibility, given $x_n$, are drawn from $
\prod\limits_{d=1}^{D}
\prod\limits_{m}^{} 
p(y_{nm}^{d},v_n^d)$. 

Note that the hierarchical mixture density in Eqn. \eqref{yDistribution} and the joint distribution in Eqn. \eqref{jointDistribution} are conditioned on $x_n$. 
All model parameters are in a functional form of $x_n$ and the joint distribution in Eqn. \eqref{jointDistribution} is differentiable. We choose to learn these functions by a CNN and the distribution is put in the loss function of the CNN. As shown in \figref{mainfigure2}, the input of the CNN is an image $x_n$ and the outputs are the HMDN parameters: $w_n^d, \mu_n^d, \sigma_n^d, \epsilon_{nj}^d, s_{nj}^d, \pi_{nj}^d$, for $d=1,...,D$ and $j=1,...,J$. The output parameters consist of three parts. $w_n^d$ is the visibility probability in Eqn. \eqref{visBernoulli}, $\mu_n^d, \sigma_n^d$ for the uni-modal Gaussian in Eqn. \eqref{uni}, and $\epsilon_{nj}^d, s_{nj}^d, \pi_{nj}^d$ for the GMM in Eqn. \eqref{gmm}. Different activation functions are used to meet the defined ranges of parameters. For instance, the standard deviations $\sigma_n^d$ and $s_{nj}^d$ are activated by an exponential function to remain positive and $\pi_{nj}^d$ by a softmax function to be in $[0,1]$.

The prediction of the visibility, the value of $w_n^d$, is used to compute the visibility loss over the visibility label $v_n^d$. See \secref{sec:traintest}. Depending on the visibility label $v_n^d$, the parameters of the uni-modal Gaussian (for visible joints) or GMM (for occluded joints) are chosen to compute the loss, as shown in blue and in orange respectively in \figref{mainfigure2}.

\begin{figure}[t]
	\centering
	\includegraphics[trim=12mm 38mm 0mm 30mm, clip, width=\linewidth]{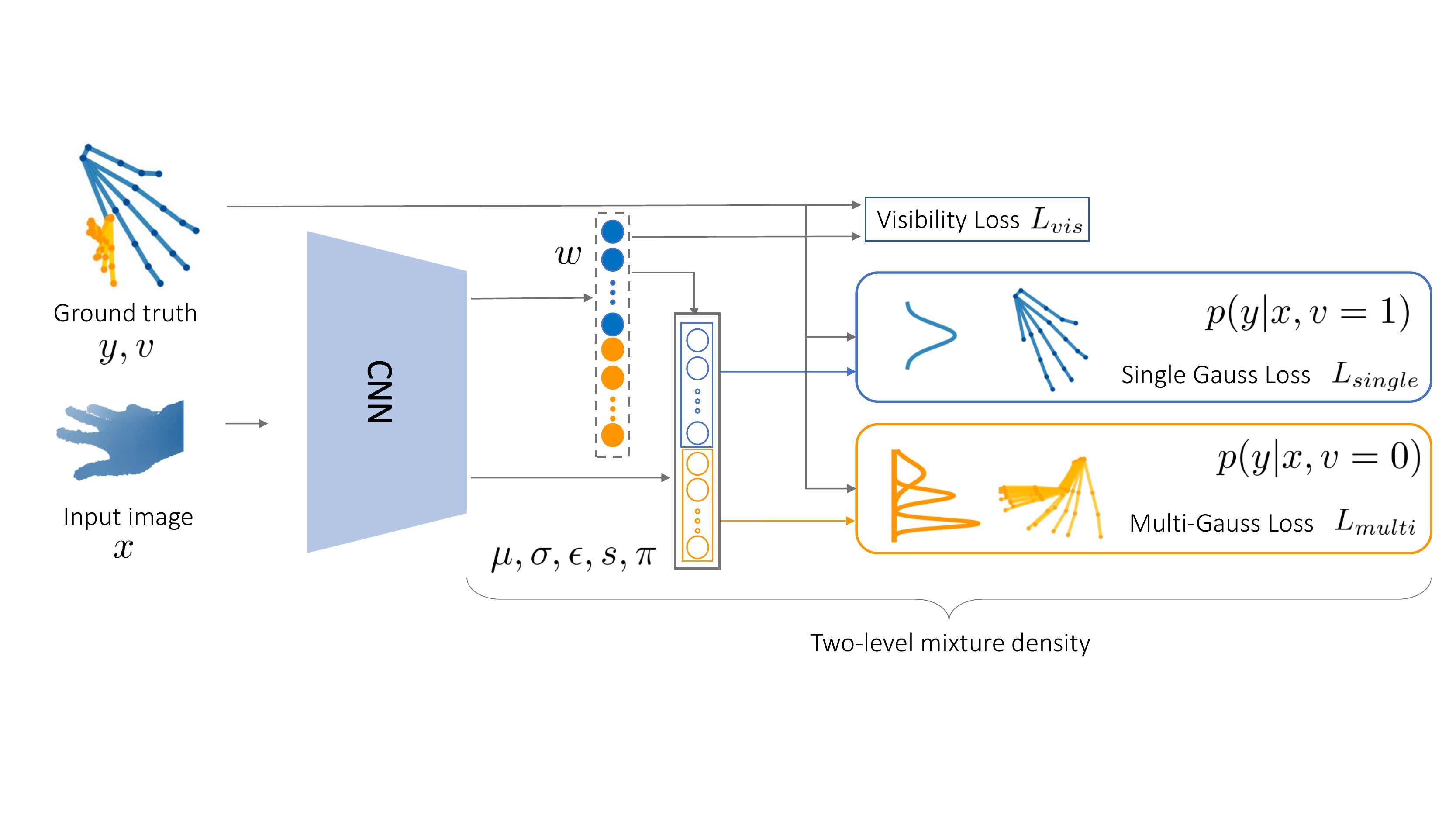}
	\caption{Hierarchical Mixture Density Network. Hand joint locations $y$ given the input image $x$ are modeled in a two-level hierarchy: in the first level, the visibility is modeled by Bernoulli distribution whose parameter is $w$; then depending on the visibility, the joint locations are either modeled by uni-modal Gaussian distributions (visible joints, shown in blue) or GMMs (occluded joint, shown in orange).	The CNN outputs the parameters of HMDN, \ie $w$, $\mu$, $\sigma$, $\epsilon$, $s$, $\pi$.}
	\figlabel{mainfigure2}
\end{figure}
%\vspace{-5mm}

\subsection{Training and Testing}
\seclabel{sec:traintest}
The likelihood for the entire dataset $\{x_n, Y_n^d, v_n^d | n=1,...,N, d=1,...,D \}$ is computed as 
$ P = \prod\limits_{n=1}^{N}\prod\limits_{d=1}^{D}\prod\limits_{m}^{} p(y_{nm}^{d},v_n^d) $, where $p(y_{nm}^{d},v_n^d)$ in \eqref{jointDistribution} has the model parameters dependent on $x_n$. Thus, our goal is to learn the neural networks that yield the parameters that maximize the likelihood on the dataset. We use the negative logarithmic likelihood as the loss function, 
\be
L= -log P = \sum\limits_{n=1}^{N}\sum\limits_{d=1}^{D}\sum\limits_{m}^{}  \{L_{vis} + L_{single} + L_{multi}\},
\eqlabel{loss}
\ee
where
\be
L_{vis} = -v_n^d log(w_n^d) - (1-v_n^d) log(1-w_n^d),
\eqlabel{vis}
\ee
\be
L_{single}= -v_n^d log(\gauss(y_{nm}^{d};\mu_n^d,\sigma_n^d)),
\eqlabel{single}
\ee
\be
L_{multi} = - (1-v_n^d) log(\sum\limits_{\gausscomp=1}^{\gausscompNum} \pi_{nj}^d \gauss(y_{nm}^{d};\epsilon_{nj}^d,s_{nj}^d)).
\eqlabel{multi}
\ee
The three loss functions correspond to the three branches in \figref{mainfigure2}. The visibility loss $L_{vis}$ is computed using the predicated value of $w_n^d$. When $v_n^d=1$, $L_{multi}=0$ and $L_{single}$ is calculated, and when $v_n^d=0$, vise versa.

During testing, when an image $x_n$ is fed into the network, the prediction for the $d$-th joint location is diverted to different branches according to the prediction of the visibility probability $w_n^d$. If $w_n^d$ is larger than 0.5, the prediction (or sampling) for the location is made by the uni-modal Gaussian distribution in Eqn. \eqref{uni}; otherwise, the GMM in Eqn. \eqref{gmm}.

However, when the prediction for the visibility is erroneous, the prediction for the joint location will be wrong. 
% the visibility variable is drawn from the predicted distribution at testing, which is different from that in training.
% as during training,  the optimization of the parameters of the uni-modal Gaussian distribution has not seen samples for occluded joints and also the optimization for the parameters of GMM has not accounted samples for visible joints. 
To help the bias problem, instead of using the binary visibility labels $v_n^d$ to compute the likelihood, we use the samples drawn from the estimated distribution in Eqn. \eqref{visBernoulli} during training. When the number of samples is large enough, the mean of these samples becomes $w_n^d$. So, the losses in Eqn. \eqref{single} and \eqref{multi} change to
\be
L_{single}= -w_n^d log(\gauss(y_{nm}^{d};\mu_n^d,\sigma_n^d)),
\eqlabel{newsingle}
\ee
\be
L_{multi} = - (1-w_n^d) log(\sum\limits_{\gausscomp=1}^{\gausscompNum} \pi_{nj}^d \gauss(y_{nm}^{d};\epsilon_{nj}^d,s_{nj}^d)).
\eqlabel{newmulti}
\ee
The modified losses in Eqn. \eqref{newsingle} and \eqref{newmulti} can be seen as a soft version of the original ones Eqn. \eqref{single} and \eqref{multi}. 
%[iterative? or how to determine w in advance?]

% for the visibility variable $v$. Therefore, the $\vistarget_\handpart^\imgIdx$ in \eqref{single}\eqref{multi} is replaced by $\sum\limits_{q=1}^{Q}{\vistarget '}_{\handpart m}^\imgIdx$, which equates $\visbleProb$ when Q is large enough. 

\subsection{Degradation into Mixture Density Network}
\seclabel{mdn}

HMDN degrades into Mixture Density Network (MDN), without the supervision for learning the visibility variable. The other form of \eqref{yDistribution} is 
\be
p(y_{nm}^d | w_n^d) = w_n^d\gauss(y_{nm}^{d};\mu_n^d,\sigma_n^d) + (1-w_n^d)\left[
\sum\limits_{\gausscomp=1}^{\gausscompNum} \pi_{nj}^d \gauss(y_{nm}^{d};\epsilon_{nj}^d,s_{nj}^d)
\right]
\eqlabel{ydist2}
\ee
where the visibility probability $w_n^d$ is learned with visibility labels. When the labels are not available, the above equation becomes 
\be
p(y_{nm}^d)= \sum\limits_{\gausscomp=1}^{\gausscompNum+1} \bar{\pi}_{nj}^d \gauss(y_{nm}^{d};\bar{\epsilon}_{nj}^d,\bar{s}_{nj}^d)
\eqlabel{degrade}
\ee 
where $\bar{\pi}_{nJ+1}^d = w_n^d, \bar{\epsilon}_{nJ+1}^d = \mu_n^d, \bar{s}_{nJ+1}^d = \sigma_n^d$, and $\bar{\pi}_{nj}^d = (1-w_n^d)\pi_{nj}^d, 
\bar{\epsilon}_{nj}^d = \epsilon_{nj}^d,
\bar{s}_{nj}^d = s_{nj}^d
$ for $j=1,...,J$.  The visibility probability $w_n^d$ in \eqref{ydist2} is absorbed into the GMM mixing coefficients $\bar{\pi}_{nj}^d$, and the distribution becomes a GMM with $J+1$ components with no dependency on the visibility. 

\section{Experiments}
\subsection{Datasets}
\seclabel{sec:dataset}
Public benchmarks for hand pose estimation are mostly collected  in third-person viewpoints and do not offer plenty of occluded joints with multiple pose labels. We investigate four datasets, ICVL~\cite{Tang_CVPR_2014}, NYU~\cite{Tompson_TOG_2014}, MSHD~\cite{Sharp_CHI_2015} and BigHand~\cite{Yuan_CVPR_2017}, and exploit those containing a higher portion of occluded joints in the following experiments. The rate of occluded finger joints and the total number of training and testing images are listed in \tableref{datasets}.

The images in these datasets are paired with pose labels i.e. joint locations, without the visibility information of the finger joints. As explained in \secref{sec:hmdn}, we use the sphere model to generate the visibility labels for training HMDN. 

%and then evaluating the results w.r.t. visible and occlued joints separately.

%\vspace{-8mm}
\begin {table}[t]
\caption {The rate of occluded finger joints and the total number of frames}
\vspace{-3mm}
\tablelabel{datasets}
\begin{center}
\begin{tabular}{  p{3.0cm}  p{2.0cm} p{2.0cm}  p{2.0cm} p{2.0cm}}
\hlineB{2}
Dataset & ICVL & NYU & MSHD & EgoBigHand \\
\hline
Train (rate/total no.) & 0.06 / 16,008 &  0.09 / 72,757 &  0.33 / 100,000 & 0.48 / 969,600 \\
Test (rate/total no.) &  0.01 / 1,596 & 0.36 / 8,252 &  0.16 / 2,000 &  0.24 / 33,468\\
\hlineB{2}
\end{tabular}
\end{center}
\end {table}
%\vspace{-8mm}

The BigHand dataset consists of two subparts: the egocentric subset includes lots of self-occlusions but lacks diverse articulations; the third-person viewpoint subset spans the full articulation space while the proportion of occluded joints, especially severe occlusions, is low. We augment the egocentric subset using the articulations of the third-person view dataset, and use it called EgoBigHand for experiments. EgoBigHand includes 8 subjects: frames of 7 subjects are used for training and frames of 1 subject for testing. 
%Cross-validation results are provided in the supplementary material. \textcolor{red}{I'd like to delete this cross-validation in case we are not able to provide. }

% The validation set consists of frames of 1 subject and randomly selected clips of frames (each clip lasts to 10 seconds) captured from 7 other subjects. The remaining frames of the 7 subjects form the training set. The testing set consists of frames (with occlusions?) from the last subject. [where else do we explain the use of the validation set? or may we explain the train/test sets only here? also, people typically request cross-validation here, may we say it is in the supplementary?]  

More results are also shown on MSHD and NYU datasets. 

\subsection{Comparison with baselines}

In the previous section, we showed that HMDN degrades to Mixture Density Network (MDN), when there is no visibility label available in training. To compare MDN with HMDN fairly, the number of Gaussian components of MDN is set same as HDMN. The other baseline is Single Gaussian Network (SGN), which is the CNN trained with a uni-modal Gaussian distribution. In \cite{PRML}, it is shown that maximization of the likelihood function under a uni-modal Gaussian distribution for a linear model is equivalent to minimizing the mean squared error errors. In our experiments, we observed that the estimation error of SGN using the Gaussian center is about the same as that of the CNN trained with the mean squared error. For further comparisons under the probabilistic framework, we report the accuracies of SGN. 
%for visible joints, the estimation error of uni-modal Gaussian model is 24.7mm and the deterministic CNN 26.5mm; for occluded joints, the estimation errors of two models are 25.4mm and 25.5mm. 

The CNN network used is the U-net proposed in \cite{Ronneberger_ICMI_2015}, by adapting the final layers to fully connected layers for regression. All the networks are trained using Adam \cite{Kingma_ICLR_2014} and the convergence times of all methods above took about 12 hours using Geforce GTX 1080Ti.\\

\begin{figure*}[t]
\centering
\includegraphics[trim=60mm 155mm 75mm 180mm, clip, width=\linewidth]{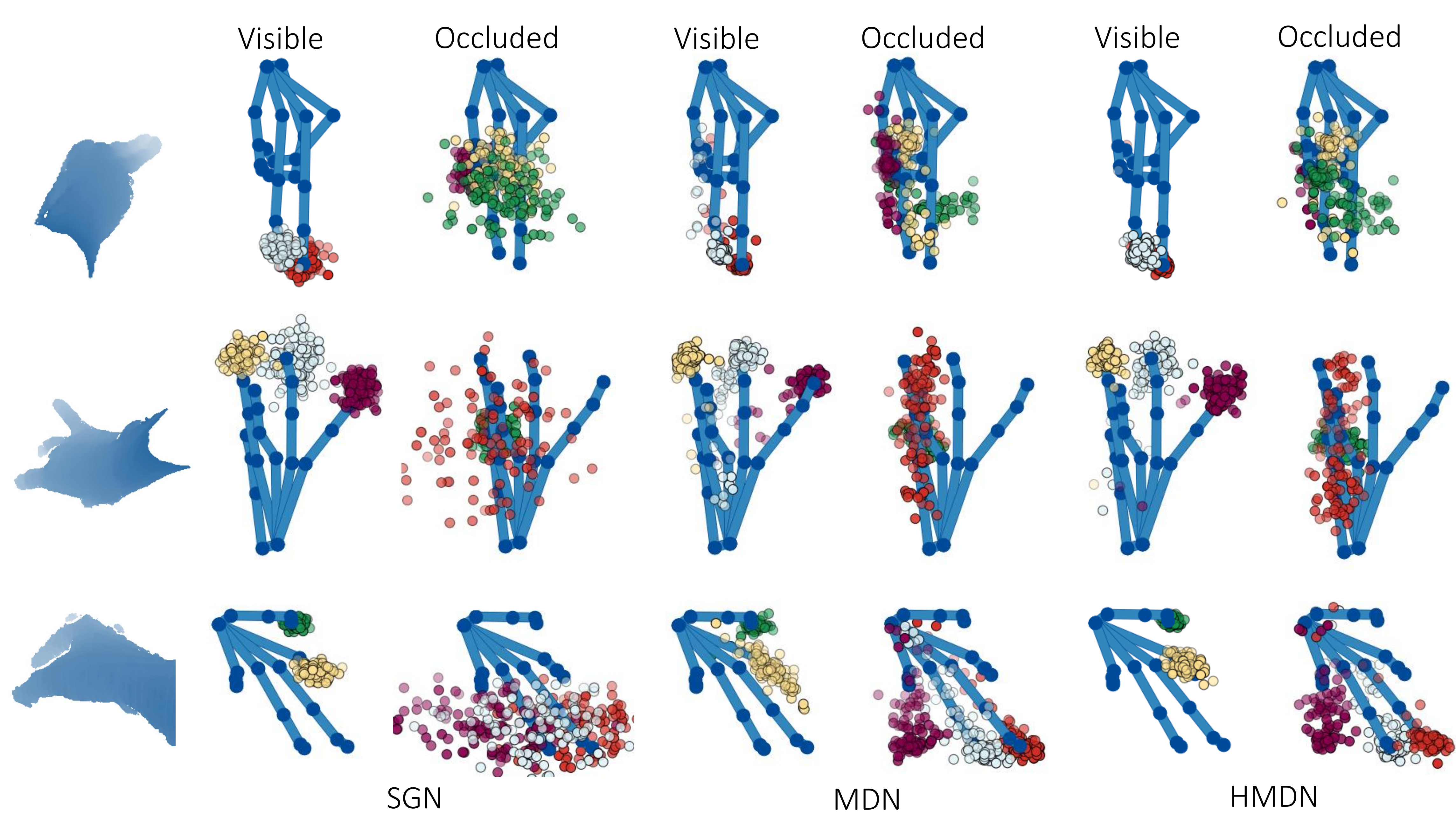}
\caption{Samples drawn from the distributions of SGN, MDN and HMDN for finger tips, shown in comparison to a pose label.}
\figlabel{qualiModels}
\end{figure*}
%\subsubsection{Qualitative Analyses}
%\seclabel{sec:QualiCmp}

\noindent \textbf{Qualitative Analyses.} See \figref{qualiModels}. 100 samples for each finger tip are drawn from the distributions of the different methods. HMDN is motivated by the intrinsic mapping difference: single-valued mapping for visible and multi-valued mapping for occluded joints. Our results, shown in \figref{qualiModels}, demonstrate its ability of modeling this difference by producing interpretable and diverse candidate samples accordingly. For visible joints, SGN and HMDN produce the samples distributed in a compact region around the ground truth location, while the samples from MDN scatter in a larger area. For occluded joints, while the samples produced by SGN scatter in a broad sphere range, the samples produced by HMDN form an arc-shaped region, which indicates the movement range of finger tips within the kinematic constraints. 

With the aid of visibility supervision, HMDN handles well the self-occlusion problem by tailoring different density functions to the respective cases. The resulting compact distributions that fit both visible joints and occluded joints improve the pose prediction accuracies in the following quantitative analyses. Such compact and interpretable distributions are also helpful for hybrid methods \cite{Tang_ICCV_2015,Sharp_CHI_2015}. For the discriminative-generative pipelines, the distribution largely reduces the space to be explored and produces diverse candidates to avoid being stuck at local minima in the generative part. For hand tracking methods \cite{Oikonomidis_BMVC_11}, the distributions of occluded joints can be combined with the motion information e.g. speed and direction, to give a sharper i.e. more confident response at a certain location.\\ 
%It is also useful for applications using reinforcement learning to speed up the learning process.
% ...
% difficulty for later stages, generative methods for examples, due to the self-similar appearance of fingers.
% ...

%\vspace{-5mm}

\begin {table}[t]
\caption {Estimation errors of different models. *see text for the evaluation metric used.}
\vspace{-3mm}
\tablelabel{quantiErrorTable}
\begin{center}
\begin{tabular}{ P{2.7cm}P{1.2cm} P{1.2cm} P{1.2cm}  P{1.2cm} P{1.2cm} P{1.2cm}P{1.2cm}}
%\begin{tabular}{ c|c|c c  |cc |cc}
\hlineB{2}
No. of Gauss.($J$) & \multicolumn{1}{c}{1}   & \multicolumn{2}{c}{10} & \multicolumn{2}{c}{20} & \multicolumn{2}{c}{30}\\
\hline  
Model  & SGN & MDN   & HMDN & MDN   & HMDN& MDN   & HMDN \\
 \hline
 Vis. Err.(mm) & 32.8&  32.2& 30.5& 34.0 & 30.7 & 32.6 & 30.5\\
 Occ. Err.(mm) & 36.5 &  35.4 & 34.8 &  36.4 & 34.4 & 35.6 &  34.2\\
*Occ. Err.(mm) & 38.9 &  34.8 &  34.6 &  35.1 & 34.2 &  35.0 & 34.5\\
\hlineB{2}
\end{tabular}
\end{center}
\end {table}

\begin{figure}[ht]
	\centering
 	\includegraphics[trim=0mm 0mm 40mm 70mm, clip, width=\linewidth]{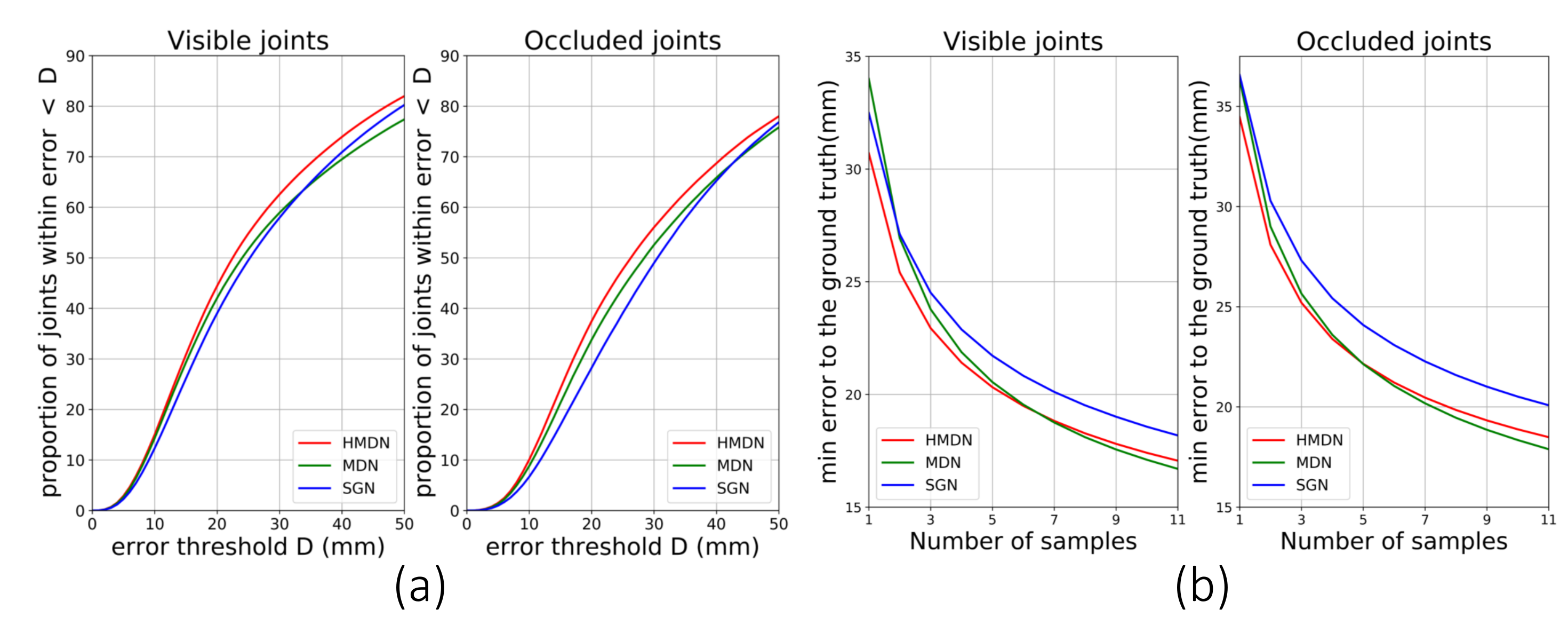}
    \caption{Comparison of HMDN, SGN, MDN, when $\gausscompNum=20$.}
	\figlabel{modelcmp}
\end{figure}

\noindent \textbf{Quantitative Analyses.} One hypothesis is drawn from the distribution of each method and is compared with the pose label, i.e. the ground truth joint location to measure the displacement error (in mm). The average errors are reported for visible joints and occluded joints separately in \tableref{quantiErrorTable}. \figref{modelcmp}a presents the comparisons under the commonly used metric, the proportion of joints within a error threshold \cite{Tang_ICCV_2015,Ye_ECCV_2016,Sharp_CHI_2015}, using 20 Gaussian components in MDN/HMDN. HMDN outperforms both MDN and SGN for visible and occluded joints using the different numbers of Gaussian components. For occluded joints, HMDN improves SGN by 10\% in the percentage of joints within the error 20mm (\figref{modelcmp}a), and by about 2mm in the mean displacement error (\tableref{quantiErrorTable}). 
HMDN also outperforms the baselines for visible joints. One can reason that given the limited network capacity, by specifying density functions by data types, HMDN learns to take a better balance between the visible and occluded, while maximizing the likelihood of the entire training data. As shown in \tableref{quantiErrorTable}, the estimation errors of HMDN do not change much for $\gausscompNum=10, 20, 30$. Note, however, the number of model parameters linearly increases with $\gausscompNum$.

% it is expected HMDN should achieve the similar error as SGN  while in \tableref{quantiErrorTable} and \figref{modelcmperrcurve}, 

% This should be for the reason that when a better distribution model is used to model the occluded joints, the optimization for the occluded joints becomes easier and the network can allocate the power to optimize the loss for visible joints. 

%Though HMDN focuses on modeling the multiple label cases, when only one label is provided, HMDN can achieve better performance than SGN.

In \figref{modelcmp}b, we vary the number of samples drawn from the distributions, and measure the minimum distance error.
HMDN consistently achieves lower errors than SGN at all numbers of samples. Compared to MDN, HMDN appears better at the smaller numbers of samples. When the number of samples increases, the error gap between the two methods becomes small.
% and then the error becomes lower than that of HMDN, which implies that the distribution of MDN has a larger variance than HMDN.

In both \tableref{quantiErrorTable} and \figref{modelcmp}, we repeated the sampling process 100 times and reported the mean accuracies. The standard deviations were fairly small as: 0.03-0.04 mm for occluded joints, and 0.01-0.02 mm for visible joints.

% In \secref{sec:hmdn}, we assume that an image with occlusions has multiple pose labels while in reality, due to noise in collecting the dataset and small hand movement, it is trivial to group pose labels for images with slight difference. During training, there is such concern: each pose label paired with its corresponding image is fed into CNN. All existing work considers images captured at different time are different and each image only has one pose label. We follow this practice and use the conventional euclidean error as our evaluation metric. 

As our motivation is in modeling the distribution of joint locations, we measure how well the predicted distribution aligns with the target distribution. As shown in \figref{ImgMultGT}, multiple pose labels are gathered for the same image with occlusions. %As in the training (\secref{sec:dataset} [it is Sec 3.1]), we form up the multiple pose lables by gathering same testing images under occlusions. 
We draw multiple samples from the predicted distribution and measure the minimum distance between the set of drawn samples and the set of pose labels. As shown in the last row of \tableref{quantiErrorTable}, the improvement is significant. Both MDN and HMDN outperform SGN by about 4 mm, which demonstrates that the arc-shaped distributions produced by MDN/HMDN align better with the target joint locations than the sphere-shaped distribution produced by SGN, as shown in \figref{qualiModels}. Instead of the minimum distance, we could use other similarity measures between distributions.\\

\begin {table}[t]
\caption {Comparison of HMDN\textsubscript{hard} and HMDN\textsubscript{soft}}
\vspace{-3mm}
\tablelabel{bias}
\begin{center}
\begin{tabular}{  P{2.7cm} P{1.5cm} P{1.5cm}  P{1.5cm} P{1.5cm} P{1.5cm} P{1.5cm}}
\hlineB{2}
No. of Gauss.($J$)   & \multicolumn{2}{c}{10} & \multicolumn{2}{c}{20} & \multicolumn{2}{c}{30}\\
\hline  
Model & HMDN\textsubscript{hard}   & HMDN\textsubscript{soft} & HMDN\textsubscript{hard}   & HMDN\textsubscript{soft}& HMDN\textsubscript{hard}   & HMDN\textsubscript{soft} \\
 \hline
 Vis. Err.(mm) &  32.2 & 30.5 & 32.9 & 30.7 &  33.1 & 30.5\\
 Occ. Err.(mm) & 35.8 & 34.8 & 35.9 &  34.4 & 36.4 & 34.2 \\
\hlineB{2}
\end{tabular}
\end{center}
\end {table}

%\subsubsection{Bias}

\noindent \textbf{Bias.} In \secref{sec:traintest}, we proposed to mitigate the exposed bias during testing, by sampling from the visibility distribution at training. HMDN trained with the loss functions in Eqn. \eqref{single} and \eqref{multi}, is denoted as HMDN\textsubscript{hard}, while the one trained with Eqn. \eqref{newsingle} and \eqref{newmulti} is HMDN\textsubscript{soft}. In \tableref{bias} HMDN\textsubscript{soft} consistently achieves lower errors than HMDN\textsubscript{hard} for different numbers of Gaussian components.

%\vspace{-5mm}
\begin{figure}[t]
	\centering
	\begin{subfigure}[t]{0.33\linewidth}	
	\includegraphics[trim=5mm 0mm 10mm 5mm, clip, width=\linewidth]{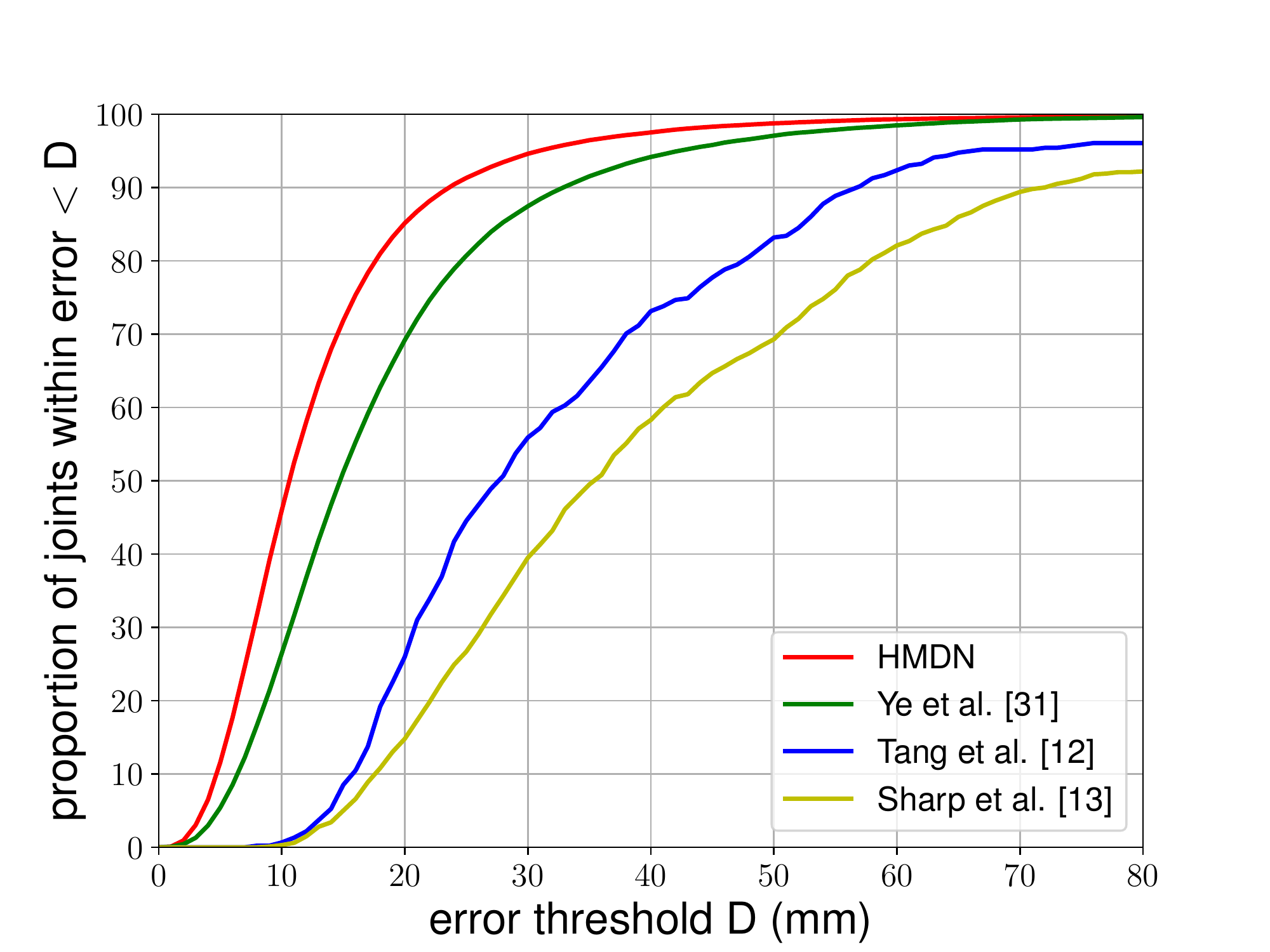}
	  \caption{}
	  \figlabel{msrcprop}		
	\end{subfigure}
	\begin{subfigure}[t]{0.32\linewidth}
	\includegraphics[trim=10mm 0mm 10mm 5mm, clip, width=\linewidth]{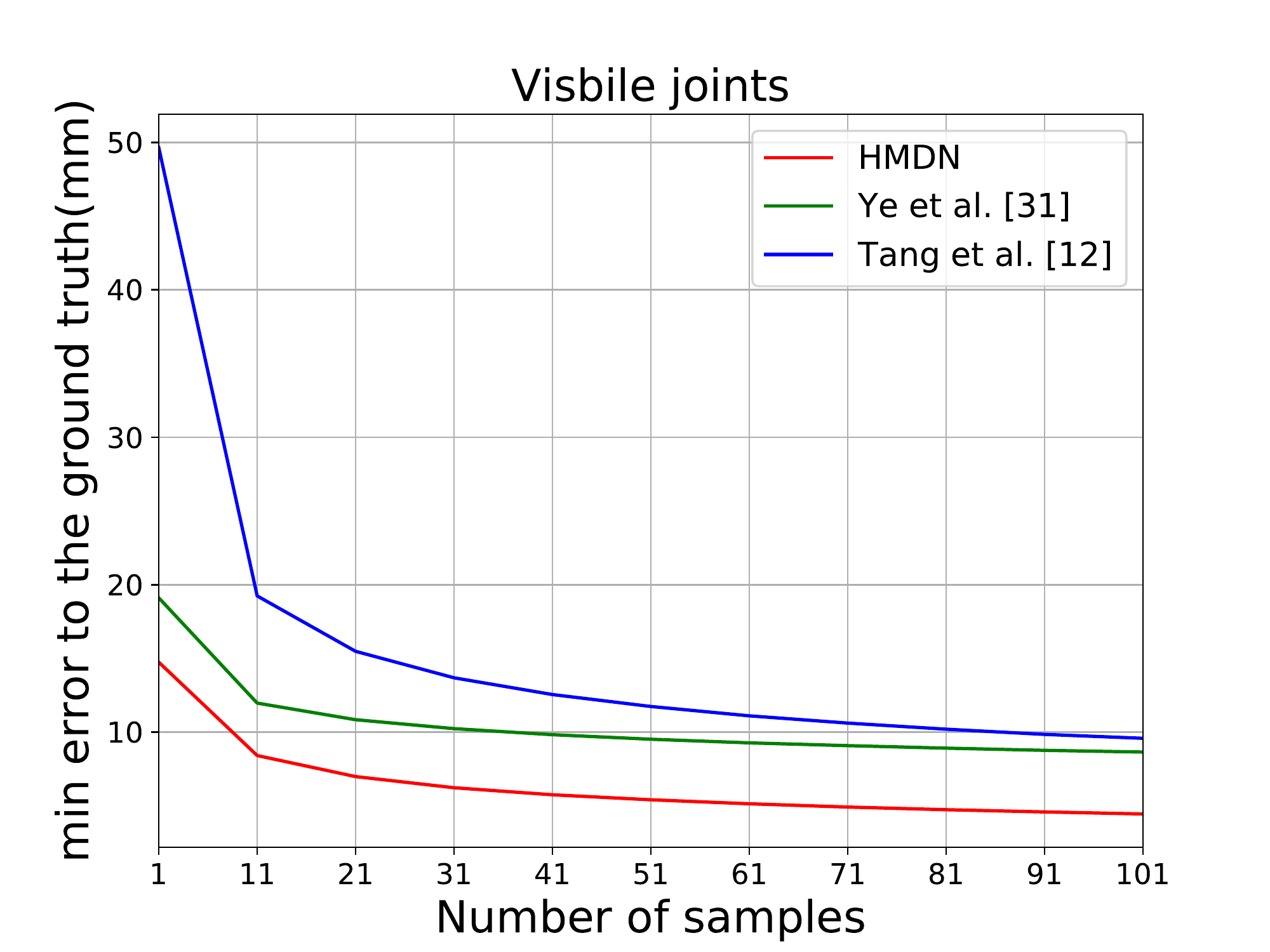}
	  \caption{}
	  \figlabel{msrcvis}		
	\end{subfigure}
	\begin{subfigure}[t]{0.32\linewidth}
	\includegraphics[trim=10mm 0mm 10mm 5mm, clip, width=\linewidth]{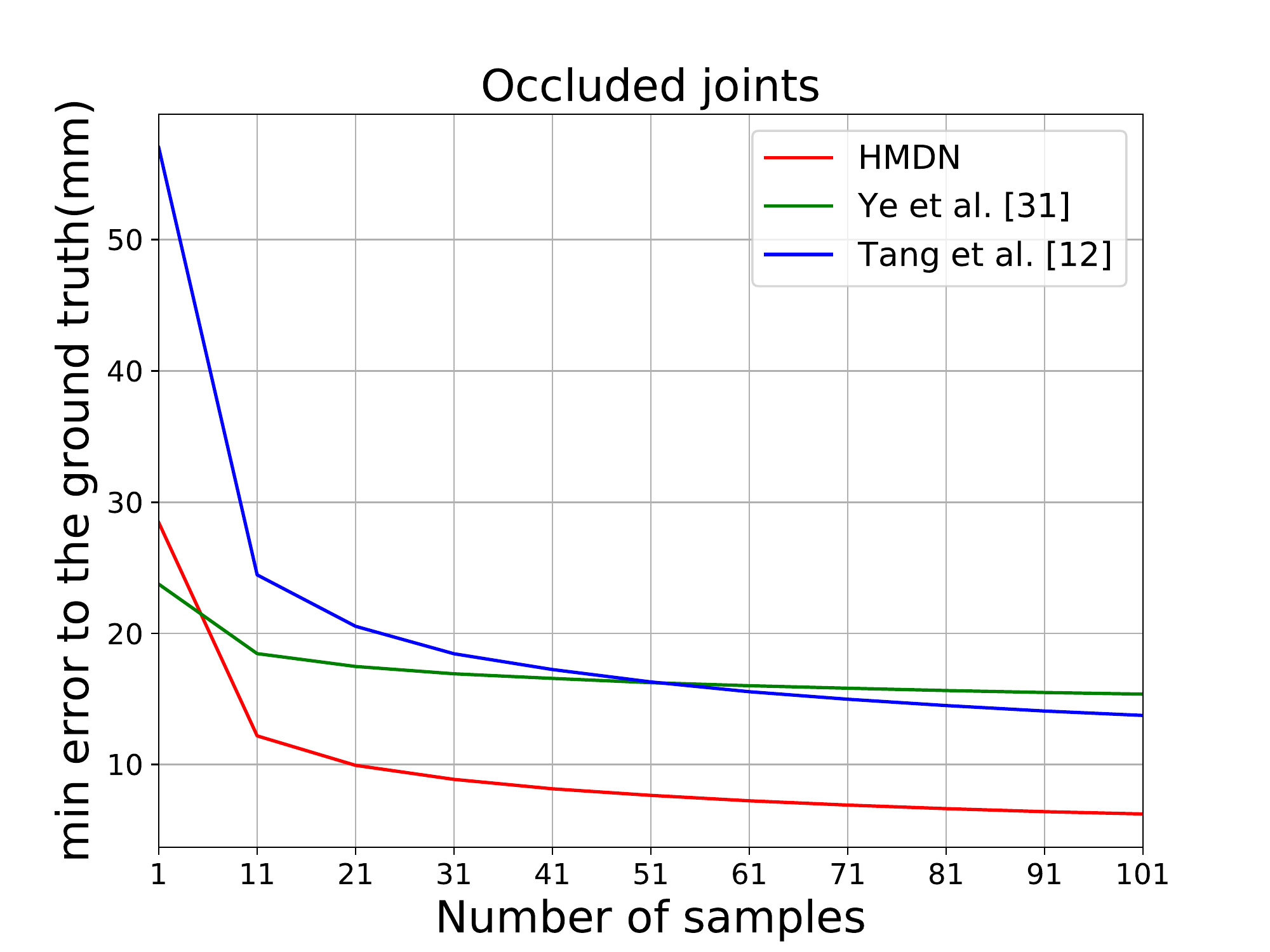}
	  \caption{}
	  \figlabel{msrcocc}	
	\end{subfigure}
	\caption{Comparison of HMDN with prior work.}
	\figlabel{msrccmp}
\end{figure}

\begin{figure}[t]
\centering
\includegraphics[trim=10mm 140mm 35mm 180mm, clip, width=\linewidth]{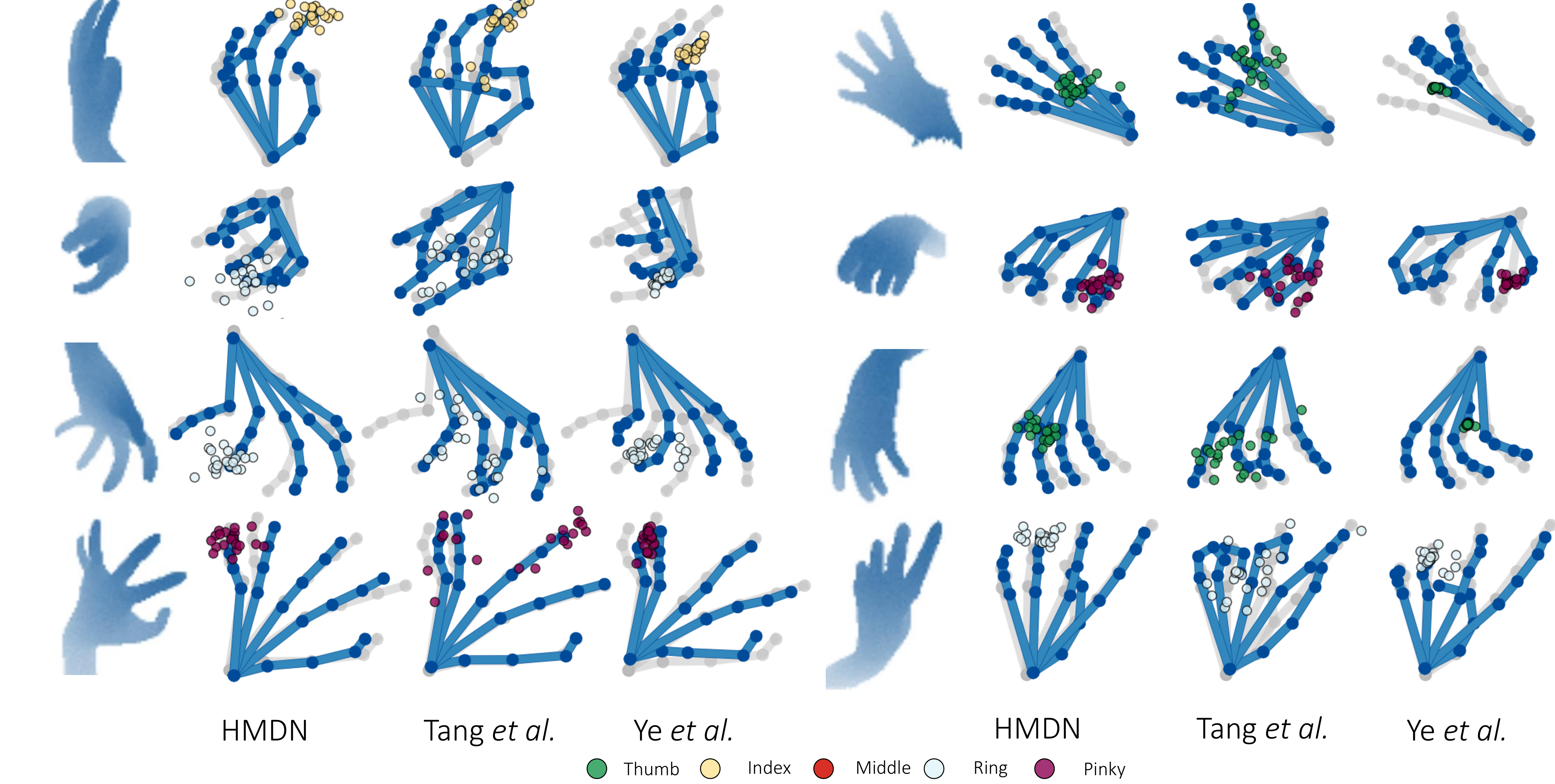}
\caption{Comparison of HMDN with Tang \etal \cite{Tang_ICCV_2015} and Ye \etal \cite{Ye_ECCV_2016}. Ground truth: skeletons in gray. Predictions from the models: skeletons in blue. For each image, samples for one tip joint from the three methods are scattered along the skeletons. Visible joints in the left column and occluded joints in the right column. }
\figlabel{msrceg}
\end{figure}
\subsection{Comparison with the state-of-the-arts}
To compete with state-of-the-arts, the following strategies are adopted: first, a CNN network is trained to estimate the global rotation and translation, and conditioned on the estimation, HMDN is then trained; data augmentation, including translation, in-plane rotation, and scaling is used.\\

%\subsubsection{MSHD Dataset} 

\noindent \textbf{MSHD Dataset.} MSHD has a considerable number of occluded joints both in training and testing set. We compare HMDN with three methods: Ye et al.\cite{Ye_ECCV_2016}, Tang et al.\cite{Tang_ICCV_2015}, Sharp et al.\cite{Sharp_CHI_2015}. For \cite{Sharp_CHI_2015}, the results of its discriminative part are used. \figref{msrcprop} shows the proportion of joints within different error thresholds for the four methods, where a single prediction is used from HMDN. 

%HMDN achieves the best performance. 
%The prediction of HMDN is the mean of the Gaussian component centres
% \footnote{If the joint is predicted to be visible, the mean of the Gaussian kernel for visible joints are chosen; if occluded, the mean of the maximum modality( the Gaussian kernel with a maximum ratio of $\mixcoeff_\gausscomp / \varo_\gausscomp $) is chosen.}.
%... it is similar to sampling a single hyphothesis as in the prior expeirments...

%3 GMMs from 3 trees, which can be seen as a GMM with 9 kernels. 
%These three methods combine discriminative approach and the generative approach and the discriminative parts are able to produce multiple hypotheses. 

In \figref{msrcvis} and \figref{msrcocc}, we further compare Ye \etal \cite{Ye_ECCV_2016} and Tang \etal \cite{Tang_ICCV_2015} with HMDN, by varying the number of hypotheses i.e. samples from the output distributions, and measuring the minimum displacement errors. Ye \etal \cite{Ye_ECCV_2016} use a deterministic CNN. To produce multiple samples, they jitter around the CNN prediction, which can be treated as a uni-modal Gaussian. Tang \etal \cite{Tang_ICCV_2015} use decision forests (3 trees) and the data points in the leaf nodes are modeled by GMM with 3 components. During testing, samples are drawn from GMMs of all trees. We used the original codes from the authors in our experiments.

HMDN significantly outperforms both methods for visible joints. For occluded joints, when the number of samples is 1, the errors of HMDN and Ye \etal\cite{Ye_ECCV_2016} are close. However, Ye \etal\cite{Ye_ECCV_2016} are not able to produce diverse samples to reach low errors as HMDN when the number of samples increases. Tang \etal\cite{Tang_ICCV_2015} provide diverse candidates by GMM in its leaf nodes, but the variance of the distribution is much larger than that of Ye \etal\cite{Ye_ECCV_2016} and HMDN for both visible and occluded joints. From the results, HMDN demonstrates its superiority for both the unimodal Gaussian model and GMM: the compact distribution with lower bias for visible joints and the diverse samples yet having smaller variances for occluded joints. See \figref{msrceg} for example results. The samples from Tang \etal \cite{Tang_ICCV_2015} for the finger tips spans a large region; those from Ye \etal \cite{Ye_ECCV_2016} are more compact but many deviate from the ground truth.\\

\begin{figure}[t]
	\centering
	\includegraphics[trim=30mm 0mm 35mm 5mm, clip, width=\linewidth]{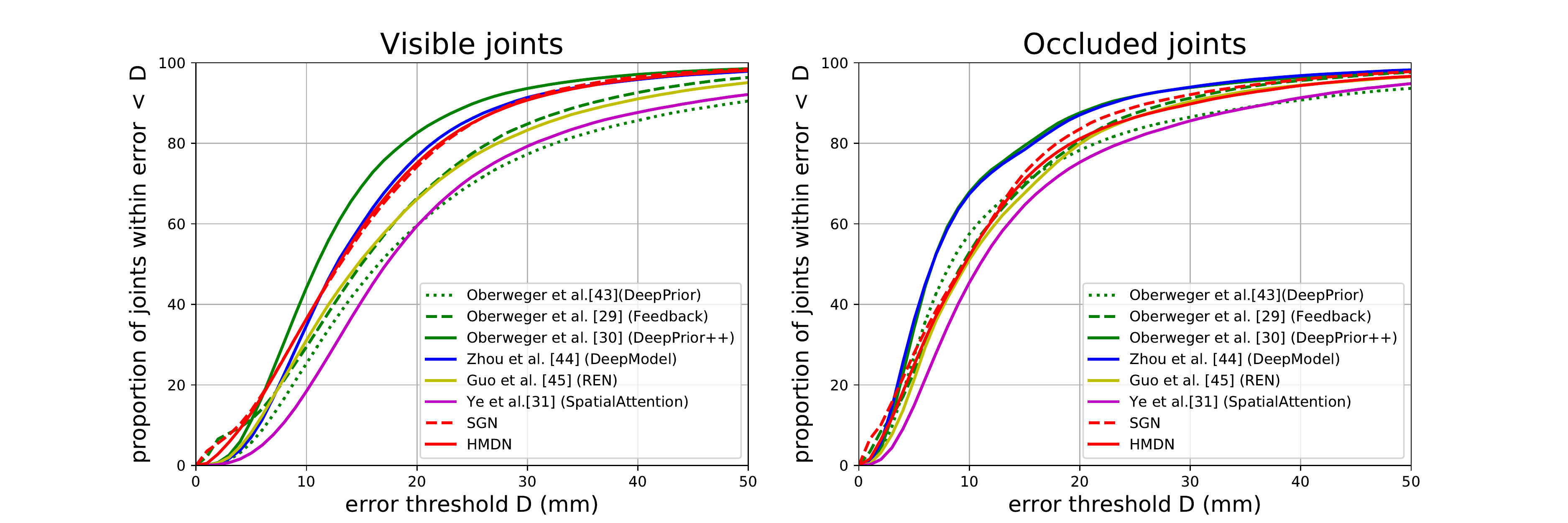}
	\caption{Comparison with state-of-the-art approaches on NYU dataset.}
	\figlabel{nyucmp}
\end{figure}

%\subsubsection{NYU Dataset}
\noindent \textbf{NYU Dataset.} The proposed method has also been evaluated on NYU dataset. Most joints in the training set are visible while on the testing set,
there are up to 36\% occluded joints. This implies all the joints in the testing dataset will be predicted as visible joints. Despite the ill-setting for HMDN, the method does not fail but degrades into SGN: the performances of SGN and HMDN are similar as shown in \figref{nyucmp}, and when compared with various state-of-the-arts based on CNN \cite{Oberweger_CVWW_2015,Oberweger_ICCV_2015,Oberweger_ICCVW_2017,Zhou_IJCAI_2016,Guo_ICIP_2016,Ye_ECCV_2016}, HMDN is in the second place for visible joints and third place for occluded joints. Note the best method \cite{Oberweger_ICCVW_2017} uses a 50-layer ResNet model \cite{He_CVPR_2016} and 21 more CNN models to refine the estimation. 
%where our HMDN only uses two CNN models.

%\textcolor{red}{ do not forget to change the reference numbers in the figures at end}

\section{Conclusion}

This paper addresses the occlusion issues in 3D hand pose estimation. 
Existing discriminative methods are not aware of the multiple modes of occluded joints and thus do not adequately handle the self-occlusions frequently encountered in egocentric views. The proposed HMDN models the hand pose in a two-level hierarchy to explain visible joints and occluded joints by their uni-modal and multi-modal traits respectively. The experimental results show that HMDN successfully captures the distributions of visible and occluded joints, and significantly outperforms prior work in terms of hand pose estimation accuracy. HMDN also produces interpretable and diverse candidate samples, which is useful for hybrid pose estimation, tracking, or multi-stage pose estimation, which require sampling. As future work, we consider modeling hand structural information i.e. finger joint dependency. This way, the sampling will produce more kinematically valid poses. Testifying HMDN on hand-object and hand-hand interaction scenarios is interesting. Though it was tested on the datasets with self-occlusions, the generalization to different occlusion types is promising.

%[are we really okay in treating joints independently in the experiments??]

%the improvement by arranging the joints in a tree structure and modelling the distribution for the joint in deeper levels conditioned on the joints in the lower levels. 

\clearpage

\bibliographystyle{splncs}
\bibliography{egbib}
\end{document}